\documentclass[letterpaper, 10 pt, conference]{ieeeconf}  

\IEEEoverridecommandlockouts                              

\def\cc{\mathbf{c} }

\def\RRR{\mathbb{R} }

\usepackage{cite}
\usepackage{amsmath,amssymb,amsfonts}
\usepackage{algorithm}
\usepackage{algpseudocode}
\usepackage{graphicx}
\usepackage{textcomp}
\usepackage[utf8]{inputenc}
\usepackage[centering,includeheadfoot,text={6.5in,9in}]{geometry}
\usepackage{xcolor}
\usepackage{subcaption}
\begin{document}

\title{\LARGE \bf
DiversityGAN: Diversity-Aware Vehicle Motion Prediction via Latent Semantic Sampling
}

\author{Xin Huang$^{1,2}$, Stephen G. McGill$^{1}$, Jonathan A. DeCastro$^{1}$,\\ Luke Fletcher$^{1}$, John J. Leonard$^{1,2}$, Brian C. Williams$^{2}$, Guy Rosman$^{1}$
\thanks{$^{1}$Toyota Research Institute, Cambridge, MA 02139, USA 
}%
\thanks{$^{2}$Computer Science and Artificial Intelligence Laboratory, Massachusetts Institute of Technology, Cambridge, MA 01239, USA
        {\tt\small xhuang@csail.mit.edu }}%
}

\maketitle

\begin{abstract}
Vehicle trajectory prediction is crucial for autonomous driving and advanced driver assistant systems. 
While existing approaches may sample from a predicted distribution of vehicle trajectories, they lack the ability to explore it -- a key ability for evaluating safety from a planning and verification perspective. 
In this work, we devise a novel approach for generating realistic and diverse vehicle trajectories.
We extend the generative adversarial network (GAN) framework with a low-dimensional \textit{approximate semantic} space, and shape that space to capture semantics such as merging and turning.
We sample from this space in a way that mimics the predicted distribution, but allows us to control coverage of semantically distinct outcomes.
We validate our approach on a publicly available dataset and show results that achieve state-of-the-art prediction performance, while providing improved coverage of the space of predicted trajectory semantics.
\end{abstract}


\vspace{-1mm}
\section{Introduction}
Vehicle trajectory prediction is crucial for autonomous driving and advanced driver assistant systems. 
For planning and verification, an ideal predictor must both accurately mimic the distribution of future trajectories and efficiently cover possible outcomes with little computational cost.
While many recent efforts cater to accuracy \cite{houenou2013vehicle,wiest2012probabilistic,gupta2018social, huang2019uncertainty,cui2018multimodal,alahi2016social}, and some work addresses diversity and coverage  \cite{hong2018diversity,gangwani2018learning,cohen2019diverse,phan2019covernet}, sampling efficiency still remains an challenge. This is important for vehicle trajectory prediction, since reasoning about the implications of predictions (e.g.\ with collision checking) is expensive \cite{schmerling2016evaluating}.

Hybrid and discrete semantics have been crucial when reasoning about agent behavior. They manifest both in representations for planners\cite{li2008generative} and in the design of tests for driving systems~\cite{esterle_specifications_2019}.   Semantic-level reasoning is crucial for attaining sample coverage across a diverse set of behaviors, but is seldom used in prediction and verification; a discrete representation may limit the expressive power of the predictor, and defining a complete taxonomy for driver trajectories is difficult, especially when reasoning about multiple agents. 

\begin{figure}[t!]
    \centering
    \begin{minipage}{0.9\linewidth}
    \includegraphics[width=1.0\linewidth]{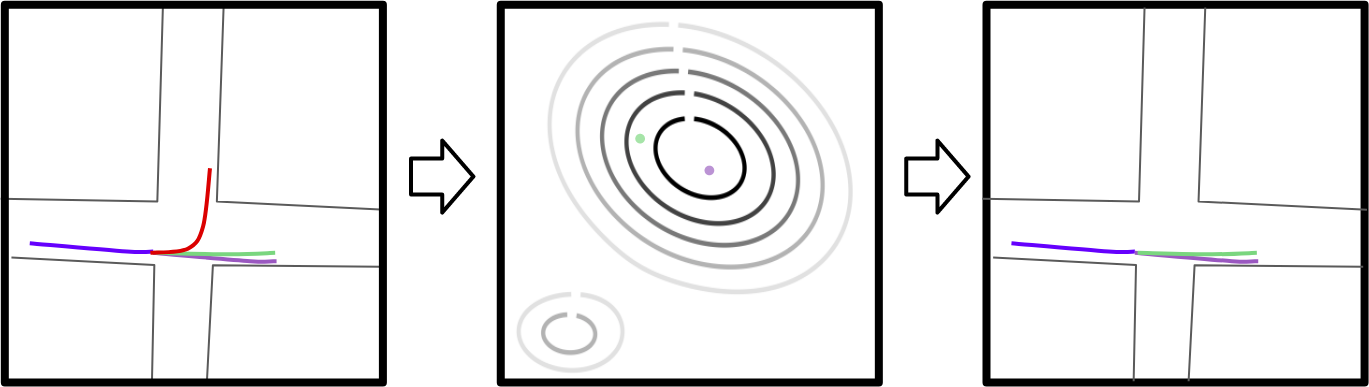}
    \end{minipage}
    
    \vspace{0.2cm}
    Direct sampling

    \vspace{0.2cm}
    \begin{minipage}{0.9\linewidth}
    \includegraphics[width=1.0\linewidth]{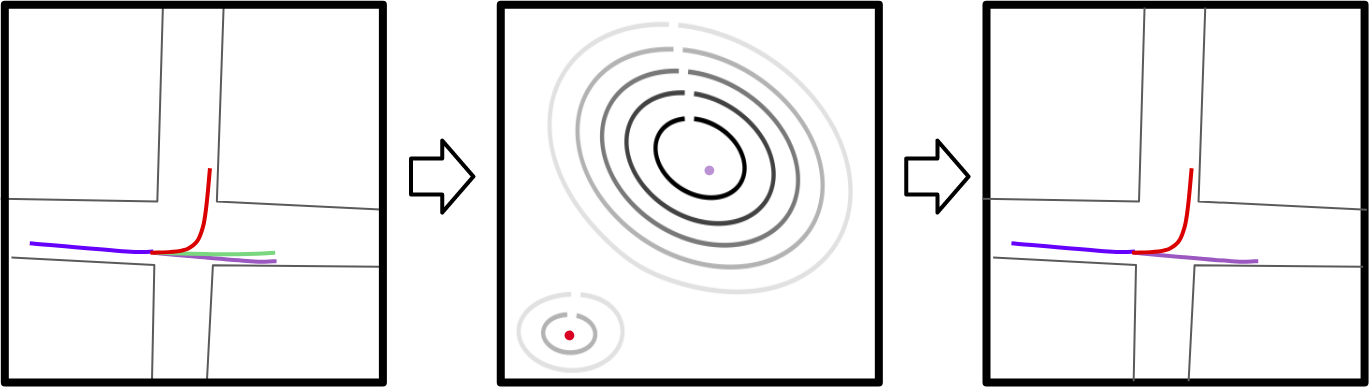}
    \end{minipage}
    
    \vspace{0.2cm}
    Latent semantic sampling

    \caption{When performing direct sampling, samples are taken uniformly from the distribution (middle), which fails to cover diverse behaviors. In latent semantic sampling, representative samples are taken, with weights associated with the distribution. In this way, a few samples can capture relevant semantic aspects, while ensuring consistency with the true prediction distribution.}
    \label{fig:latent_model}
\vspace{-5mm}
\end{figure}

In this paper, we propose a model that is both accurate and diverse by incorporating a latent semantic layer into the trajectory generation. This layer represents high-level vehicle behaviors, matching discrete semantic outcomes when they are defined. Our construction of that layer does not require a complete taxonomy of maneuvers, and can extend easily to multiple definitions of semantics such as interaction types.
We expect driving semantics to have a low effective dimensionality at every instance, since a driver can perform only a few distinct maneuvers at any given moment. 
We illustrate our approach in Figure~\ref{fig:latent_model}, where the goal is to produce diverse trajectory predictions and cover distinct outcomes.
The top row shows traditional sampling, which fails to sample diverse behaviors efficiently. The bottom row demonstrates our latent semantic sampling technique, which is able to capture both maneuvers that can be performed in the intersection. 

We avoid the need for a taxonomy by shaping the intermediate layer via metric learning \cite{weinberger2009distance}.
We train the latent semantic layer activations to match annotations of high-level labels when they exist.
Distances in the semantic layer between two trajectories should be large if they represent different semantic labels, and should
be small otherwise.

Finally, our proposed latent state affords some interpretation of the network, which is crucial in safety-critical tasks, such as autonomous driving. By tuning the high-level latent vector, our samples better cover the human intuition about diverse outcomes. 

Our work has three main contributions: i) We extend a generative adversarial network to produce diverse and realistic future vehicle trajectories. We do so using a latent layer, which is shaped via metric learning to capture semantic context, as well as trajectory geometry.
ii) We describe an efficient sampling method to cover the possible future actions and their likelihoods, which is important for safe motion planning and realistic behavior modeling in simulation.
iii) We validate our approach on a publicly available dataset with vehicle trajectories collected in urban driving.  
Quantitative and qualitative results show our method successfully learns a semantically meaningful latent space that allows for generating diversified trajectories efficiently, while achieving state-of-the-art accuracies.


\vspace{-2mm}
\subsection{Related Work}
\label{sec:related_works}

Our work relates to several topics in probabilistic trajectory prediction. Unlike deterministic alternatives \cite{houenou2013vehicle}, it allows us to reason about the uncertainty of driver's behavior.
There are several representations that underlie reasoning about trajectories. \cite{wiest2012probabilistic,huang2019uncertainty,IvanovicPavone2019,tang2019mfp,chai2019multipath} predict future vehicle trajectories as Gaussian mixture models, whereas \cite{kim2017probabilistic} utilizes a grid-based map.
In our work, we focus on generating trajectory samples directly from an approximated distribution space, using a sequential network, similar to \cite{gupta2018social}.

For longer-term prediction horizons, additional context cues are needed from the driving environment. Spatial context, such as mapped lanes and scene information, not only indicates the possible options a vehicle may take (especially at intersections), but also improves the prediction accuracy, as vehicles usually follow lane centers closely \cite{cui2018multimodal,chang2019argoverse} or follow common movements in similar scene layouts \cite{huynh2019trajectory}. 
Another important cue is social context, which allows for reasoning about interaction among agents \cite{alahi2016social,gupta2018social,IvanovicPavone2019}.
Our method takes advantage of these two cues by feeding map data and nearby agent positions into our model, improving the accuracy of predictions over a few seconds.

Recently proposed generative adversarial networks (GANs) can sample trajectories by utilizing a vehicle trajectory generator and a discriminator that distinguishes real trajectories and trajectories produced by the generator \cite{gupta2018social,li2019conditional,amirian2019social}. Despite their success, efficiently producing unlikely events, such as lane changes and turns, remains a challenge. These events are important to consider, as they can pose a significant risk and affect driving decisions. 
\begin{figure*}[t!]
    \centering
    \includegraphics[width=0.85\linewidth]{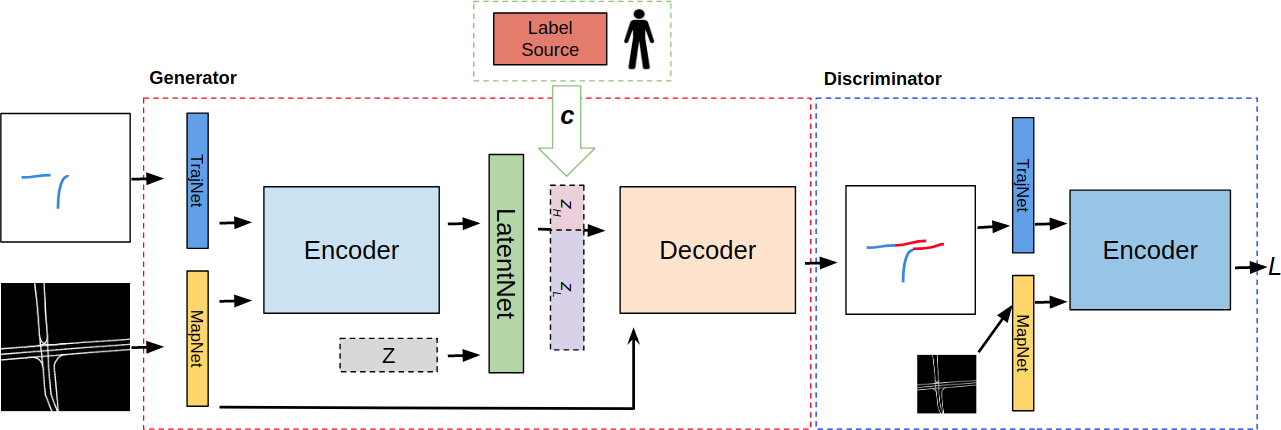}
    \caption{Architecture diagram of prediction model. We shape the space of the intermediate vector $z_H$ to resemble a human's concept of semantic distances and then use it to modify the samples that are fed to the decoder. Inputs include past vehicle trajectories and map information represented as arclength-parameterized curves (cf. text).}
    \label{fig:architecture_diagram}
\vspace{-5mm}
\end{figure*}

Hybrid models \cite{cui2018multimodal, deo2018multi} are effective at producing distinct vehicle behaviors. They classify modes (or maneuvers) first before predicting the future positions.
As such, they are restricted to cases where a fixed taxonomy of modes are well defined. In many cases, especially with multiple agents, a complete taxonomy of driving scenarios is hard to obtain. 
In \cite{phan2019covernet}, a set of trajectory modes are proposed to cover all possible motions, given vehicle dynamics. This provides sufficient coverage but 
suffers from the large number of samples predicted, as further reasoning about their implications (e.g., with collision checking) is expensive.
On the other hand, our method allows more sufficient coverage with fewer samples and handles more general cases involving with undefined semantics, including multi-vehicle interactions. 

Many recent models use an intermediate representation \cite{bengio2013representation}, to improve performance and sample efficiency \cite{bengio2013better}. \cite{tang2019mfp} utilizes a set of discrete latent variables to represent different driver intentions and behaviors. \cite{shen2019interpreting} explores the semantic features in the latent space of GANs, while InfoGAN in
\cite{chen2016infogan} successfully distills important features in the latent space based on an information bottleneck approach. In \cite{amirian2019social}, a network based on InfoGAN is proposed to produce predictions that preserve multi-modality. In addition to GANs, \cite{hristov2019disentangled} learns a disentangled latent representation in a variational autoencoder (VAE) framework to ground spatial relations between objects, while \cite{yuan2019diverse,salzmann2020trajectron++} propose a conditional VAE latent variable framework to handle multi-modality in trajectory predictions.
Unlike \cite{chen2016infogan, amirian2019social}, we use metric learning \cite{weinberger2009distance} to explicitly capture high-level notions, such as maneuvers and interactions, by training the latent space to match human similarity measures. It allows us to sample from the learned geometry to obtain distinct vehicle behaviors efficiently, without assuming the dissimilarity in trajectory space captures semantic meaning, as in \cite{yuan2019diverse}. 
In other domains, \cite{UnterthinerNSKH18} diversifies samples via a potential field motivation for image generation GANs. \cite{yang2019diversity} regularizes the generator by introducing multiple noise sources, to encourage multi-modal behaviors in computer vision tasks. \cite{cohen2019diverse} explores diversity in policy through conjugated policies with a few samples in reinforcement learning. \cite{xu2018diversity} encourages diversity in text generation by assigning higher rewards to novel outputs. 


Finally, our work has applications to sampling and estimation of rare events in support of verification tasks, which is its own active field, see \cite{bhatia_incremental_2004, bucklew2013introduction, schmerling2016evaluating, o2019scalable}
and references therein.  
Efficient sampling is crucial for safety reasoning, as verification of safety properties for a given driving strategy requires numerous simulations over a large number of scene conditions and agent behaviors.
The closest work to ours is~\cite{o2019scalable, koren2019efficient}, which also propose sample-based estimation of probabilities.  Our work focuses explicitly on sampling from diverse modes of behaviors, and effectively improves both search efficiency and representational power, allowing sufficient coverage with fewer simulations than standard Monte Carlo.

\vspace{-2.5mm}
\section{Model}
\label{sec:model}

Here, we present the problem formulation and describe the model underlying our work, including loss functions and our proposed sampling procedure. 

\vspace{-1.2mm}
\subsection{Problem Formulation}
The input to the trajectory prediction problem includes a sequence of observed vehicle trajectories $\mathbf{X} = X_1, X_2, \ldots, X_{t_{obs}}$, as well as the surrounding lanes denoted as $M$. 
Our goal is to predict a set of possible future trajectories $\mathbf{\hat{Y}} = \hat{Y}_{t_{obs}+1}, \hat{Y}_{t_{obs}+2}, \ldots, \hat{Y}_{t_{obs}+t_{pred}}$, where the observed future trajectories are denoted as $\mathbf{Y} = Y_{t_{obs}+1}, Y_{t_{obs}+2}, \ldots, Y_{t_{obs}+t_{pred}}$.

In the probabilistic setting, since multiple future trajectory sets are possible, the goal is to estimate the predicted probability distribution $P(\mathbf{Y}|\mathbf{X},M)$. As $P(\mathbf{Y}|\mathbf{X},M)$ often lacks a closed-form expression, many approaches use some form of sample generation techniques, including traditional ones such as MCMC and particle filters \cite{karasev2016intent}, planning-based approaches such as RRTs\cite{aoude2011mobile}, and probabilistic generative networks such as GANs. 

\vspace{-1.2mm}
\subsection{Model Overview}
We now describe the network structure and sampling approach, as illustrated in Figure~\ref{fig:architecture_diagram}. 
The trajectory generator takes the past trajectory of target vehicles, a map of lane centerlines, and a noise sample, to produce samples of future trajectories. 
The discriminator identifies whether the generated trajectory is realistic.

\textbf{Semantic training cues} - In addition to the generator and discriminator networks, our model assumes a supervisory source for trajectory semantics.
Both similarity votes \cite{BorgGroenen2005} and discrete label vectors \cite{strecha2011ldahash} have been used for metric learning of semantically meaningful spaces, from both human annotations, or from computational surrogates, such as classifiers.
In the context of driving, labels can include maneuvers, such as merging, turning, or slowing down, or interaction patterns such as giving right of way or turning at a four-way-stop junction.
For the purpose of our experiments, we hand-coded detectors for specific maneuvers, yielding three-logic (True/False/``undefined'') values. These are coded as vectors $\cc$ with elements $c_l \in \{-1, 1, \phi\}$.
We chose three-valued logic since in some instances no Boolean choice makes sense -- e.g., ``lane keep'' defined in \cite{deo2018multi} is ambiguous if the vehicle is driving on an open road without lane markers. 
This motivates a representation that avoids a single taxonomy of all road situations with definite semantic values.

\vspace{-1.2mm}
\subsection{Trajectory Generator}
The trajectory generator predicts realistic future vehicle trajectories, given inputs of the past trajectories and the map information. It embeds the inputs before sending them into a long short-term memory (LSTM) network encoder.
The encoder output is combined with a noise vector generated from a standard normal distribution, and fed into a latent network that separates the combined information into a high-level vector and a low-level vector.
The decoder, taking these two vectors, produces the trajectory samples.

\subsubsection{Trajectory Network}
A series of fully connected layers that embed spatial coordinates into a trajectory embedding vector \cite{alahi2016social}.

\subsubsection{Map Network}
\label{sec:map_net}
We represent nearby lanes via their polynomial coefficients, taken from the road point that is closest to the vehicle. We use arclength parameterization \cite{do2016differential} to resample the road and project it onto a second-order polynomial. 
The nearby lane coefficients are fed to the network as an input vector via a simple non-linear fully connected network, which outputs a map embedding vector. In our experiments, this improves training efficiency while maintaining accuracy, compared to rasterizing the road network (e.g., in \cite{amini2019variational,cui2018multimodal}).

\subsubsection{Encoder}
An LSTM network that encodes the spatial and map embedding vectors from time steps $1$ to $t_{obs}$ into an encoder hidden state vector.

\subsubsection{Latent Network}
The encoder hidden state is transformed, along with a standard Gaussian noise sample, into a latent state vector, $(z_H,z_L)$:  $z_H \in \RRR^{d_H}$ represents high-level information, such as maneuvers, whereas $z_L \in \RRR^{d_L}$ represents fine trajectory information. We set $d_H \ll d_L$ so that $z_H$ can be sampled efficiently.
As described in Section~\ref{sec:losses}, we encourage $z_H,z_L$ to be uncorrelated, and $z_H$ to separate semantically different trajectories.
This representation disentangles semantic concepts from low-level trajectory information, similar to information bottlenecks \cite{chen2016infogan}, but is shaped by human notions of semantic similarity, as learned from the labels.

\subsubsection{RNN-based decoder}
An LSTM network takes $z_H$, $z_L$, and a map embedding vector, and generates a sequence of future vehicle positions.

\subsection{Trajectory Discriminator}

An encoder converts the past trajectory, future predictions, and map information into a label $L$ = \{fake, real\}, where fake means a trajectory is generated by our predictor, while real means the trajectory is from data. The structure of the discriminator mirrors that of the trajectory encoder, except in its output dimensionality.

\subsection{Losses}
\label{sec:losses}
Similar to \cite{gupta2018social}, we measure the performance of our model using the average displacement error (ADE) of Equation~\ref{eq:lossade} and the final displacement error (FDE) of Equation~\ref{eq:lossfde}.
\begin{align}
\mathcal{L}_{\textit{ADE}}(\hat{Y}) &= \frac{1}{t_{pred}} \sum_{t=t_{obs}+1}^{t_{obs}+t_{pred}} ||Y_t - \hat{Y}_t||_2
\label{eq:lossade} \\
\mathcal{L}_{\textit{FDE}}(\hat{Y}) &= ||Y_{t_{obs}+t_{pred}} - \hat{Y}_{t_{obs}+t_{pred}}||_2
\label{eq:lossfde}
\end{align}

\subsubsection{Best prediction displacement loss}
Also as in \cite{gupta2018social}, we compute the Minimum over N (MoN) losses to encourage the model to cover groundtruth options, while maintaining diversity in its predictions:

\begin{equation}
    \mathcal{L}_{\textit{MoN}} =  \min_n\left(\mathcal{L}_{\textit{ADE}}\left(\hat{Y}^{(n)}\right)\right),
\label{eq:mon}
\end{equation}
where $\hat{Y}^{(1)}, \ldots, \hat{Y}^{(N)}$ are samples generated by our model.
The loss, over $N$ samples from the generator, is computed as the average distance between the best predicted trajectories and observed future trajectories.
Although minimizing MoN loss leads to a diluted probability density function compared to the groundtruth \cite{thiede2019analyzing}, we use it to show that our method can estimate an approximate distribution efficiently.
We defer a different, more accurate, supervisory cue to future work. 

\subsubsection{Adversarial loss}
We use standard binary cross entropy losses, $\mathcal{L}_{GAN,G}, \mathcal{L}_{GAN,D}$, to compute the loss between outputs from the discriminator and the labels.
These losses are used to encourage diversity in predictions.

\subsubsection{Latent space regularization loss}
We encourage the two latent space components $\mathbf{z}_L,\mathbf{z}_H$ to be independent and normally distributed with a unit variance for each vector element. We do so by adding the two regularization terms,

\vspace{-1em}  
{\footnotesize
\begin{equation}
\mathcal{L}_{\textbf{ind}} = \left(\sum_{i=1}^{d_H} \sum_{j=1}^{d_L} z_H^{i} z_L^{j} \right)^2, \; 
    \mathcal{L}_{\textbf{lat}} = 
    \begin{array}{l}
          \|\Sigma_{z_H}-I\|^2_F+ \|\mu_{z_H}\|^2_F +\\
          \|\Sigma_{z_L}-I\|^2_F + \|\mu_{z_L}\|^2_F
    \end{array},
\end{equation}
}
\!\!where $\|\cdot\|^2_F$ denotes the Frobenius norm, $I$ denotes the identity operator with appropriate dimensions, and the means and variances are computed as empirical estimates at each batch. 




\subsubsection{Embedding loss}
After enforcing $\mathbf{z}_H$ and $\mathbf{z}_L$ are independent vectors, we introduce an embedding loss to enforce the correlation between high-level latent vector $\mathbf{z}_H$ and prediction coding $\cc$. Similar to \cite{rosman2017hybrid}, if two data samples have the same answer element for label $l$, we expect the differences in their high-level latent vectors to be small. On the other hand, if two predictions have different codings, we want to encourage the difference to be large. Note that it would be ideal to take human votes on the similarity between trajectories, and in this work, we use $\cc$ as a surrogate to approximate their votes in order to mimic human notions of semantic similarity. 
The loss can be written as 
\begin{equation}
    \mathcal{L}_{\textbf{emb}} = \sum_{m=1,n=1}^{B}\sum_{l=1}^{s} \text{sign}\left(c_l^{(m)}, c_l^{(n)}\right) ||\mathbf{z}_H^{(m)} - \mathbf{z}_H^{(n)}||_2,
\label{eq:embedding}
\end{equation}
where $B$ is batch size, $s \leq z_H$ is the number of defined labels, $c_l^{(m)},c_l^{(n)}$ denote the label $l$ answers on examples $m,n$ respectively, and $\text{sign}(\cdot,\cdot)=0$ if either argument is $\phi$.

\subsubsection{Total loss}
In total, we combine the losses listed above together with appropriate coefficients.
{\footnotesize
\begin{align}
    \mathcal{L,D} &= \mathcal{L}_{GAN,D} \\
    \mathcal{L,G} &= \lambda_1\mathcal{L}_{\textit{MoN}} + \lambda_2\mathcal{L}_{GAN,G} + \lambda_3\mathcal{L}_{\textbf{ind}} + \lambda_4\mathcal{L}_{\textbf{lat}} + \lambda_5\mathcal{L}_{\textbf{emb}}
\label{eq:loss_all}
\end{align}}
\vspace{-8mm}

\subsection{Sampling Approach}
We now describe how we sample from the space of $z_H$ in Alg.~\ref{alg:sampling}.
We generate a set of latent samples with size $N_{all}$, selecting from them a subset of $N$ representatives using the Farthest Point Sampling (FPS) algorithm\cite{GONZALEZ1985293,Hochbaum:1985:BPH:2775965.2775967}.
We store the nearest representative identity as we compute the distances, to augment the FPS representatives with a weight proportional to their Voronoi cell weight, computed as the ratio of the number of cell samples to total sample size.
This gives us a \textit{weighted} set of samples that converges to the original distribution, but favors samples from distinct regions of space.
FPS allows us to emphasize samples that represent distinct high-level maneuvers encoded in $z_H$.

\begin{algorithm}
\caption{Semantic Sampling\label{alg_sampling}}
\begin{algorithmic}[1]
\ForAll {$i=1\ldots N_{all}$}
  \State Sample from $z_{(i)}\sim Z$.
  \State Generate latent sample $(z_{H,(i)},z_{L,(i)})$.
\EndFor
\State Perform Farthest Point Sampling on $\{\mathbf{z}_{H,(i)}\}$ to obtain $N$ representative samples, $\{(\mathbf{z}_{H,(j)},\mathbf{z}_{L,(j)})\}_{j=1}^N$
\State Compute Voronoi weights $w_j$ for each FPS sample. 
\State Decode from $(\mathbf{z}_{H,(j)},\mathbf{z}_{L,(j)})$ a full prediction $Y_{(j)}$.
\State Return $\{(Y_{(j)},w_j)\}_{j=1}^{N}$
\end{algorithmic}
\label{alg:sampling}
\end{algorithm}
The samples cover (in the sense of an $\epsilon$-covering) the space of possible high-level choices. 
The high-level latent space is shaped according to human notions of semantic similarity. With this similarity metric shaping, FPS can leverage its $2$-optimal distance coverage property in order to capture the majority of semantically different prediction roll-outs in just a few samples.\footnote{We note that a modified FPS\cite{volkov2015coresets} can trade off mode-seeking with coverage-seeking when generating samples.}



\vspace{-0.2mm}
\section{Results}
\label{sec:result}
In this section, we describe the details of our model and dataset, followed by a set of quantitative results against state-of-the-art baselines and qualitative results on accurate and diverse predictions.
\vspace{-0.2mm}
\subsection{Model Details}
The \textit{Trajectory Network} utilizes two stacked linear layers with (32, 32) neurons.
The \textit{Map Network} uses four stacked linear layers with (64, 32, 16, 32) neurons.
An LSTM with one layer and a hidden dimension of 64 forms both the \textit{Encoder} and \textit{Decoder} in the \textit{Trajectory Generator}.
The \textit{Latent Network} fuses the Encoder output and a 10-dimensional noise vector. This network is composed of two individual linear layers with output dimensions of 2 and 72 for the high-level and low-level layers, respectively.
The \textit{Discriminator} is an LSTM with the same structure as the Generator's Encoder, followed by a series of stacked linear layers with dimensions of (64, 16, 1), activated by a sigmoid layer at the end.
All linear layers in the Generator are followed by a batch norm, ReLU, and dropout layers.
The linear layers in the Discriminator use a leakyReLU activation instead.
The number of samples $n$ we use for the MoN loss is 5.
The loss coefficients in Eq.~\eqref{eq:loss_all} are selected to be 4, 1, 100, 2, 50, respectively.

The model is implemented in Pytorch and trained on a single NVIDIA Tesla V100 GPU. We use the Argoverse forecasting dataset \cite{chang2019argoverse} for training and validation, and select the trained model with the smallest MoN ADE loss on the validation set.

\vspace{-0.5mm}
\subsection{Semantic Annotations}
In order to test our embedding over a large scale dataset, we devised a set of classifiers for the data as surrogates to human annotations.
They check for specific high-level trajectory features, and each of them outputs a ternary bit representing whether the feature exists, does not exist, or is unknown.
The list of feature filters used in this paper includes: accelerate, decelerate, turn left, turn right, lane follow, and lane change.
\vspace{-0.3mm}
\subsection{Prediction Accuracy}

\begin{figure*}[t!]
\begin{subfigure}[b]{0.48\linewidth}
\centering
\includegraphics[width=0.85\columnwidth]{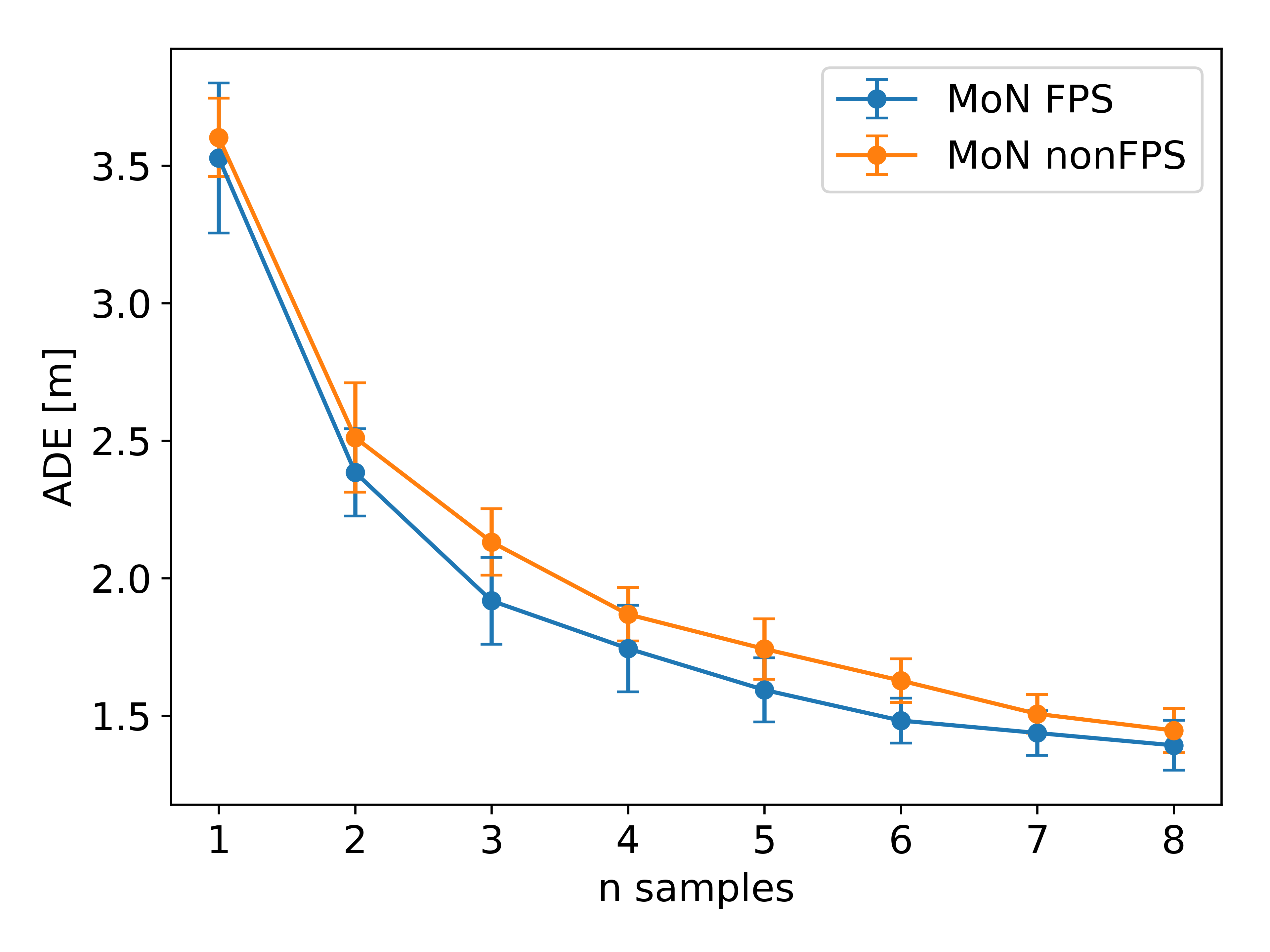}

(a)
\end{subfigure}
\begin{subfigure}[b]{0.48\linewidth}
  \centering
    \includegraphics[width=0.85\columnwidth]{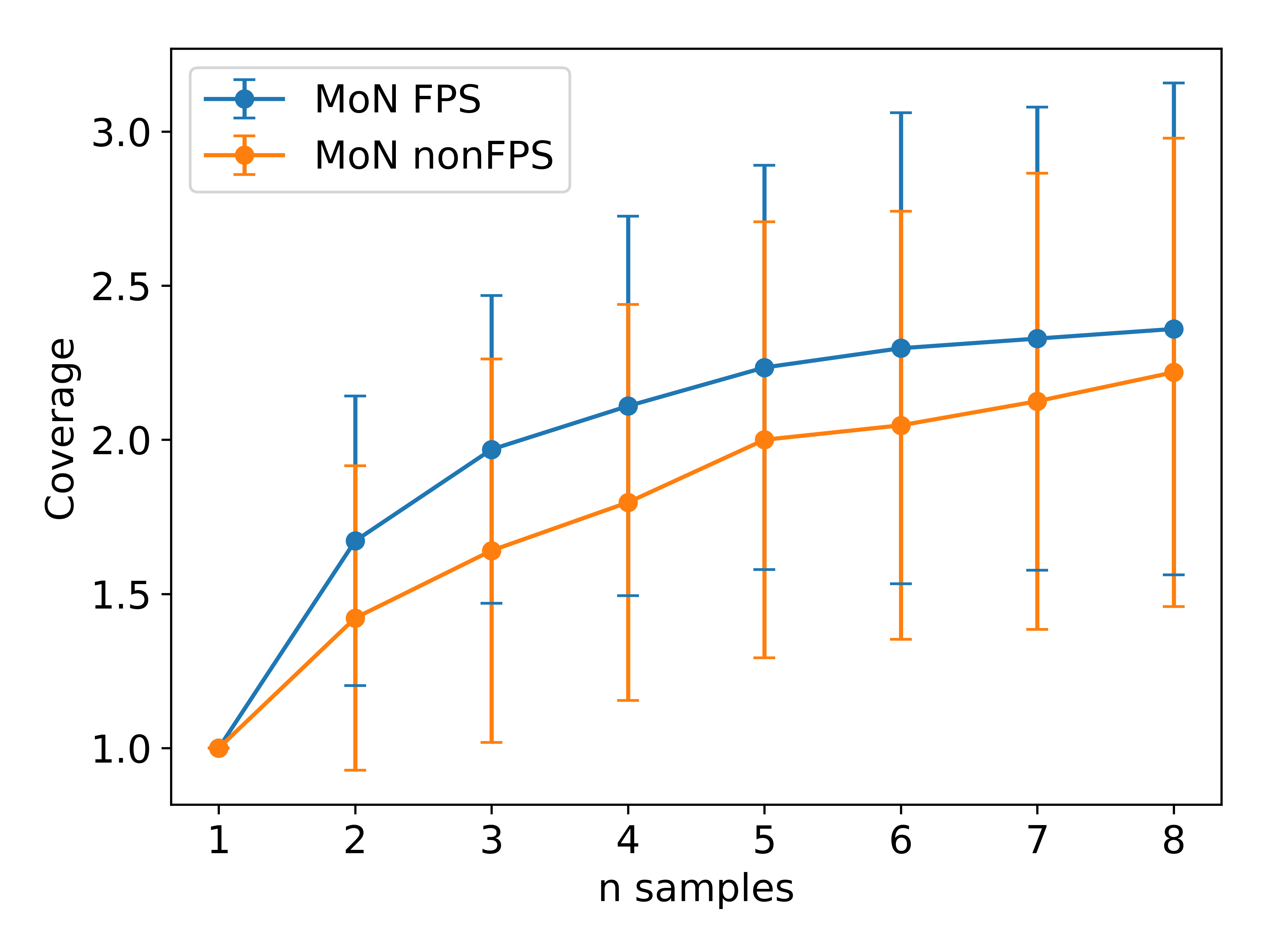}
    
    (b)
  \end{subfigure}
\caption{Accuracy and coverage comparisons between FPS sampling (blue) and direct sampling (orange) over 3 seconds with $N$ from 1 to 8. (a) Accuracy comparison using MoN as the metrics, where the gap between two curves indicates the improvement using FPS, especially when $N$ is from 2 to 6. (b) Coverage comparison, where $y$-axis measures the number of distinct discrete label codings extracted from the predicted trajectories. The gap indicates that FPS achieves better coverage of prediction options with smaller numbers of samples.
}
  \label{fig:mon_comparison}
\vspace{-3mm}
\end{figure*}

Over 1 and 3 second prediction horizons, with $N=5$ samples, we compute the MoN ADE~\eqref{eq:lossade} and FDE~\eqref{eq:lossfde} losses, respectively.
In addition to our method, we introduce a few baseline models to demonstrate the prediction accuracy of our method.
The first two baselines include a linear Kalman filter with a constant velocity (CV) model and with a constant acceleration (CA) model, respectively.
We sample multiple trajectories given smoothing uncertainties.
The third baseline is an LSTM-based encoder-decoder model \cite{chang2019argoverse}, which produces deterministic predictions.
The fourth baseline is SocialGAN \cite{gupta2018social}, based on its open-source model.
In order to show the efficacy of our approach in separating and shaping the latent space, we also introduce a vanilla GAN-based model that is similar to our network, but does not regularize the latent space using the embedding loss \eqref{eq:embedding}. The vanilla model has a few variants that take different input features, where social contains the positions of nearby agents and map contains the nearby lane information as described in \ref{sec:map_net}. 

The results are summarized in Table~\ref{table:errors}. The first two rows indicate that physics-based models can produce predictions with reasonable accuracy. Using five samples, constant velocity (CV) Kalman Filter outperforms a deterministic deep model with results shown on the third row.
The rest of the table shows that a generative adversarial network improves upon accuracy by a large margin compared to physics-based models using five samples.
It is observed that the map features contribute more to long-horizon predictions. 
Additionally, our method is competitive compared to standard ones (e.g., vanillaGANs) and outperforms an off-the-shelf SocialGAN model trained on Argoverse, after regularizing the latent space using loss functions defined in Section~\ref{sec:losses}, while adding sample diversification, as shown in the rest of the section. Our results on the Argoverse test set ranked 5th in the Argoverse competition and earned Honorable Mention in the Machine Learning for Autonomous Driving (ML4AD) workshop at Neurips 2019.


\begin{table}[t]
\centering
\begin{tabular}{lclclclcl}
          & \multicolumn{2}{l}{1 Second} & \multicolumn{2}{l}{3 Seconds} \\
Model Name & ADE           & FDE          & ADE           & FDE           \\
\hline \hline 
Kalman Filter, CV          & 0.51             & 0.79            & 1.63             & 3.62             \\
Kalman Filter, CA          & 0.69             & 1.22            & 2.87             & 7.08             \\
LSTM Encoder-Decoder          & 0.57             & 0.94            & 1.81             & 4.13             \\
SocialGAN          &  0.47            & 0.69            & 1.72             & 3.49             \\
vanillaGAN          &  0.42            & \textbf{0.62}            & 1.55             & 3.09             \\
vanillaGAN+social          & 0.44             & 0.66            & 1.68             & 3.04             \\
vanillaGAN+social+map          & 0.44             & 0.63            & 1.34             & 2.75             \\
DiversityGAN+social+map          & \textbf{0.41}             & 0.65            & 1.35             & 2.74    \\ 
DiversityGAN(FPS)+social+map          & 0.44             & \textbf{0.62}            & \textbf{1.33}             & \textbf{2.72}\\
\hline \\   
\end{tabular}
\caption{MoN average displacement errors (ADE) and final displacement errors (FDE) of our method and baseline models with $N = 5$ samples.}
\label{table:errors}
\vspace{-6mm}
\end{table}

\vspace{-0.5mm}
\subsection{Latent Space Learning}
To demonstrate the latent space is learned to be semantically meaningful, we measure the k-nearest neighbor (kNN) entropy \cite{lombardi2016nonparametric} of the label distribution in the latent space, using both our method and a vanilla GAN model without an embedding loss. 

The kNN entropy is computed as follows: First, we generate a large number of $S$ latent space samples and obtain $S'$ local neighborhoods by sub-sampling $S'$ points, and $k$-nearest-neighbors sets. We then generate the trajectories and their labels for each neighborhood, and compute the entropy of the labels, which measures how well we decouple distinct labels (low entropy indicates decoupled labels). We select $S = 1000, S' = 30, k = 40$ to provide ample coverage in the low-dimensional latent space.

In the validation dataset, our learned latent space has an average entropy of 0.55, with a standard deviation of 0.33. In contrast, the latent space from a vanilla GAN has an average entropy of 1.45, with a standard deviation of 0.29. Furthermore, on average the percentage of the majority vote within each neighborhood is 71.53\% using our method and 50.94\% using a vanilla GAN. This suggests that our method successfully shapes the latent space, so as to locally disentangle semantic space. In the remainder of this section, we complete our claim that the learned space is semantically meaningful by showing that far away samples are generating trajectories with distinct labels through FPS.

\vspace{-0.5mm}
\subsection{Latent Space Sampling}

To show the effectiveness of our latent sampling approach, we measure the MoN loss with and without the FPS method.
We test using a challenging subset of the Argoverse validation dataset that filters out straight driving with constant velocity scenarios, resulting in a trajectory distribution that emphasizes rare events in the data.
As indicated in Figure~\ref{fig:mon_comparison}(a), when the number of samples increases, the prediction loss using FPS drops faster compared to direct sampling. We note the improvement is larger in the regime of $2$-$6$ samples, where reasoning about a full roll-out of multiple hypotheses is still practical in real-time systems, and we obtain an improvement of $8\%$. Beyond the improved accuracy, the proposed method is able to sample the additional modes of the distribution of trajectories, which is validated in Figure~\ref{fig:mon_comparison}(b), where we compare the coverage of distinct maneuver codes by sampled predictions and show that FPS has better coverage with smaller numbers of samples.
We further demonstrate this advantage with a small number of samples in Section~\ref{subsec:qualitative}.



\subsection{Qualitative Examples}
\label{subsec:qualitative}
\begin{figure}[t!]
\centering
\begin{subfigure}[b]{0.48\columnwidth}
    \includegraphics[width=4cm]{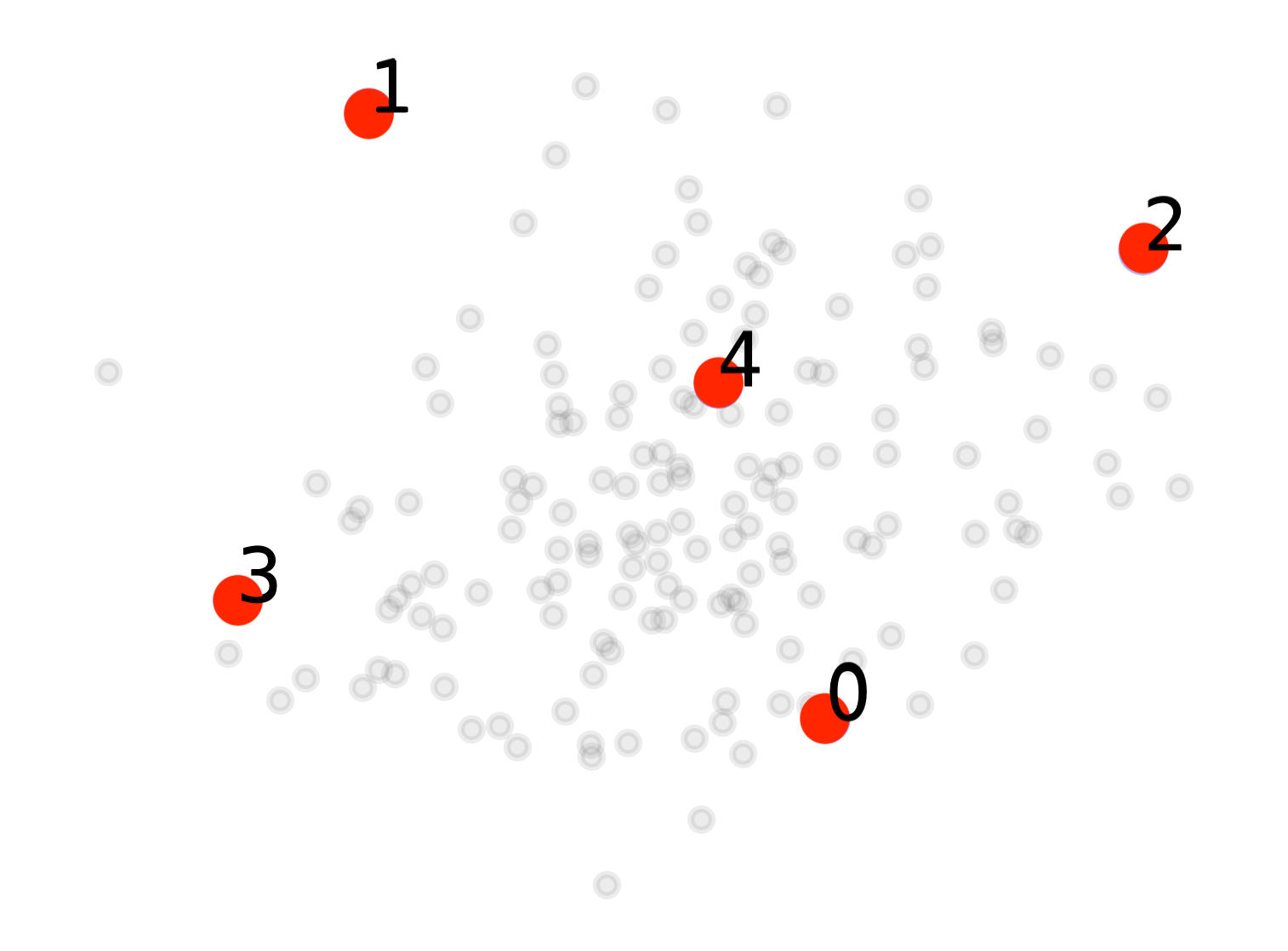}
  \end{subfigure}
    \begin{subfigure}[b]{0.48\columnwidth}
    \includegraphics[width=4cm]{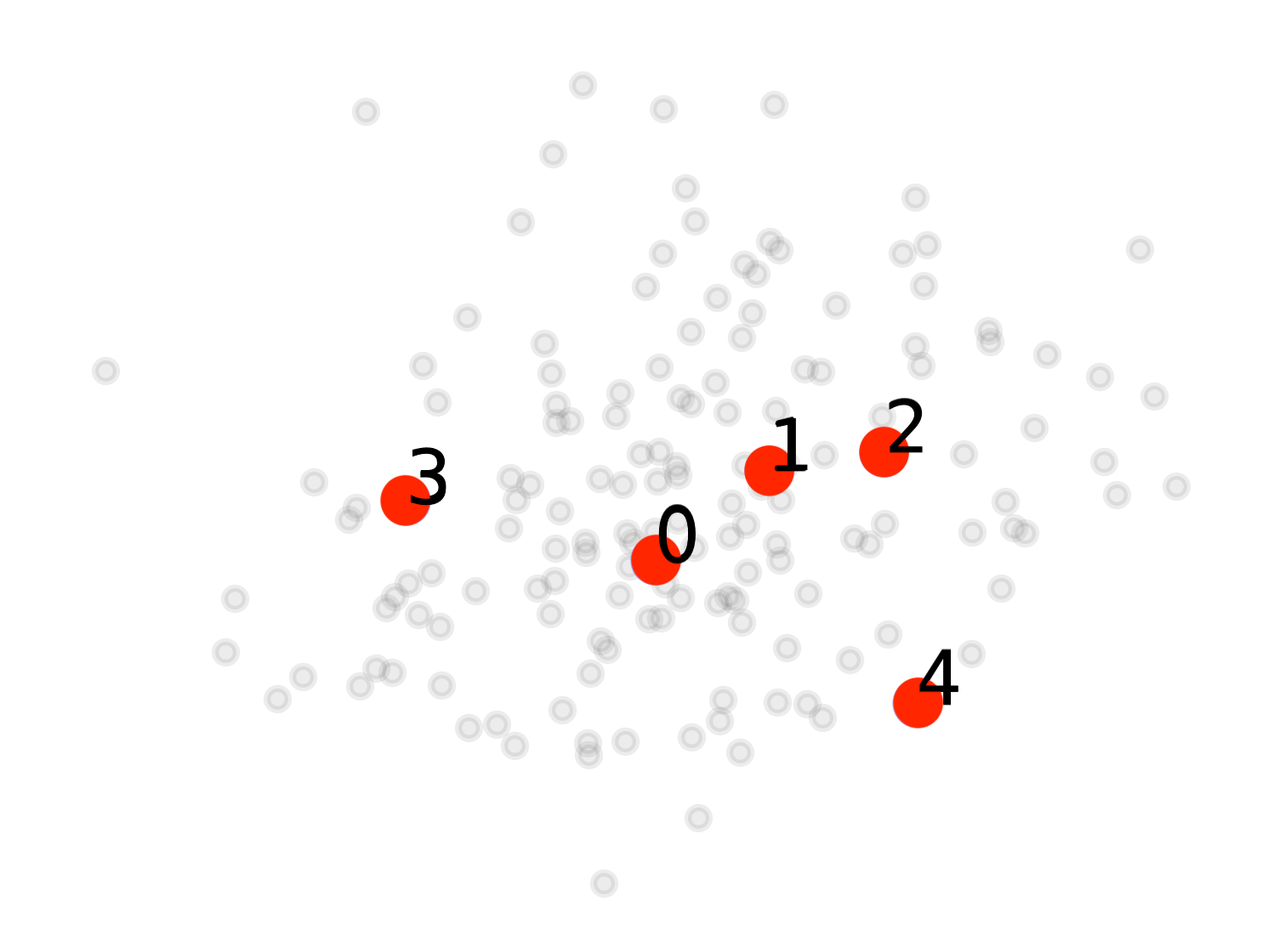}
  \end{subfigure}
  \\
\begin{subfigure}[b]{0.48\columnwidth}
    \includegraphics[width=3.6cm]{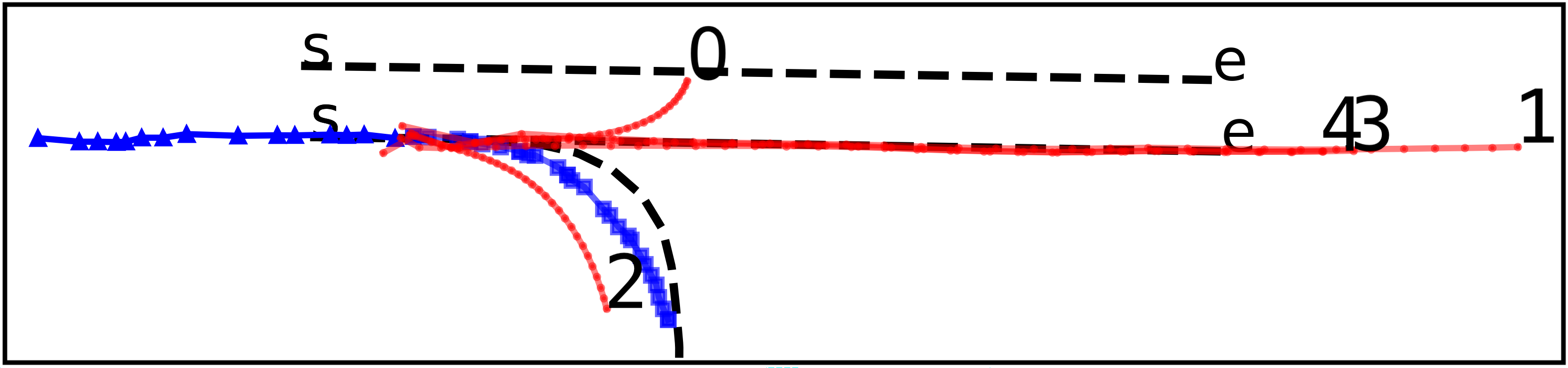}
  \end{subfigure}
  \begin{subfigure}[b]{0.48\columnwidth}
    \includegraphics[width=3.6cm]{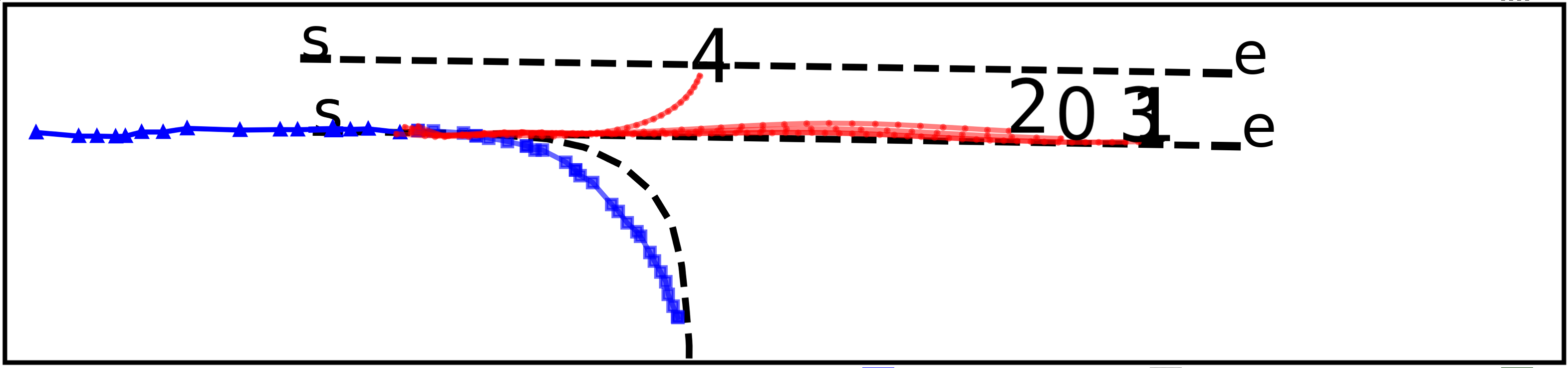}
  \end{subfigure}
  \\
  \hspace{1.3cm} FPS \hspace{2.2cm} Direct Sampling
  \\
  (a) FPS provides accurate coverage of observed future trajectory by generating rare turning samples.
  \\
  \vspace{1mm}
  \begin{subfigure}[b]{0.48\columnwidth}
    \includegraphics[width=4cm]{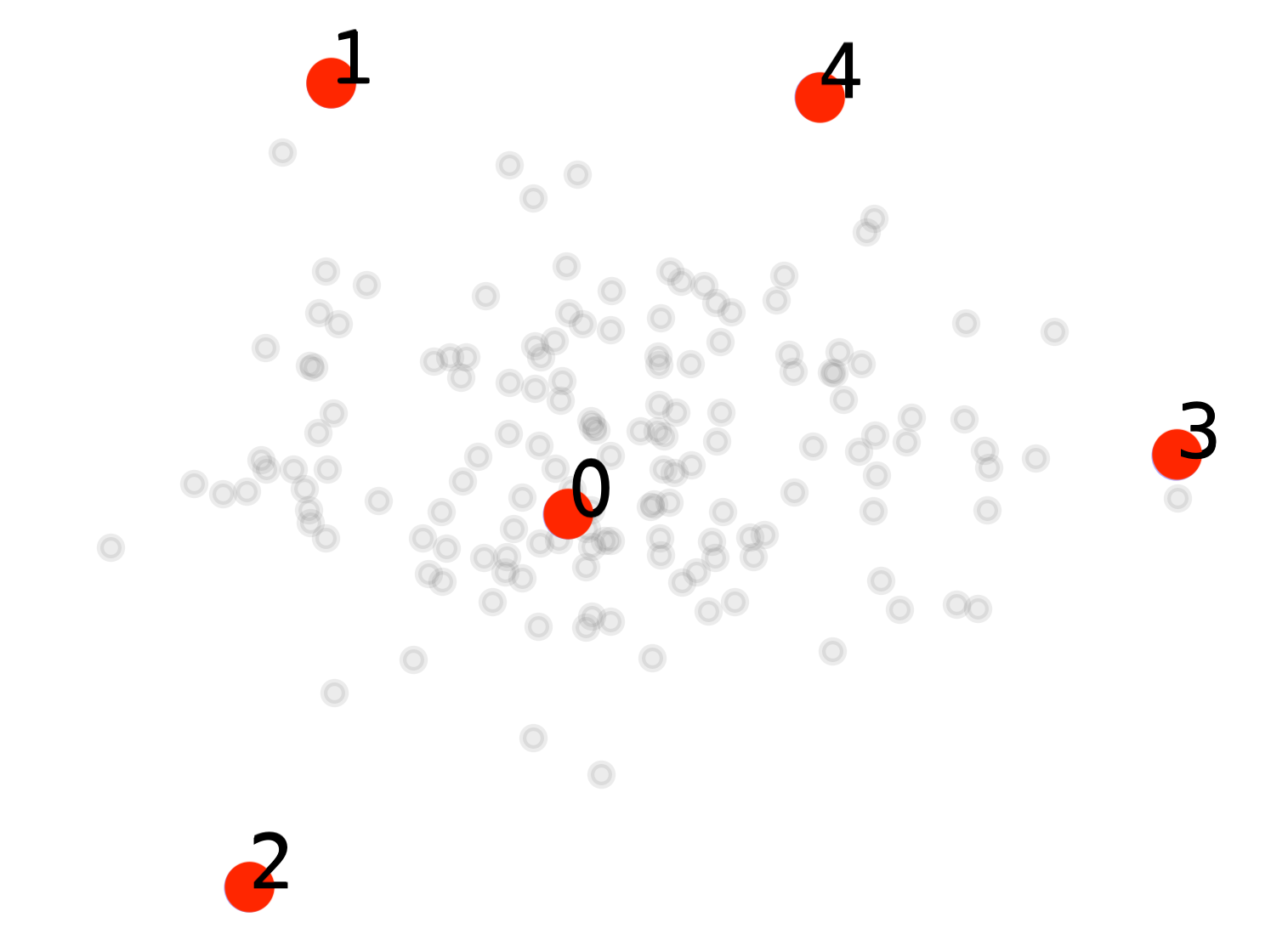}
  \end{subfigure}
    \begin{subfigure}[b]{0.48\columnwidth}
    \includegraphics[width=4cm]{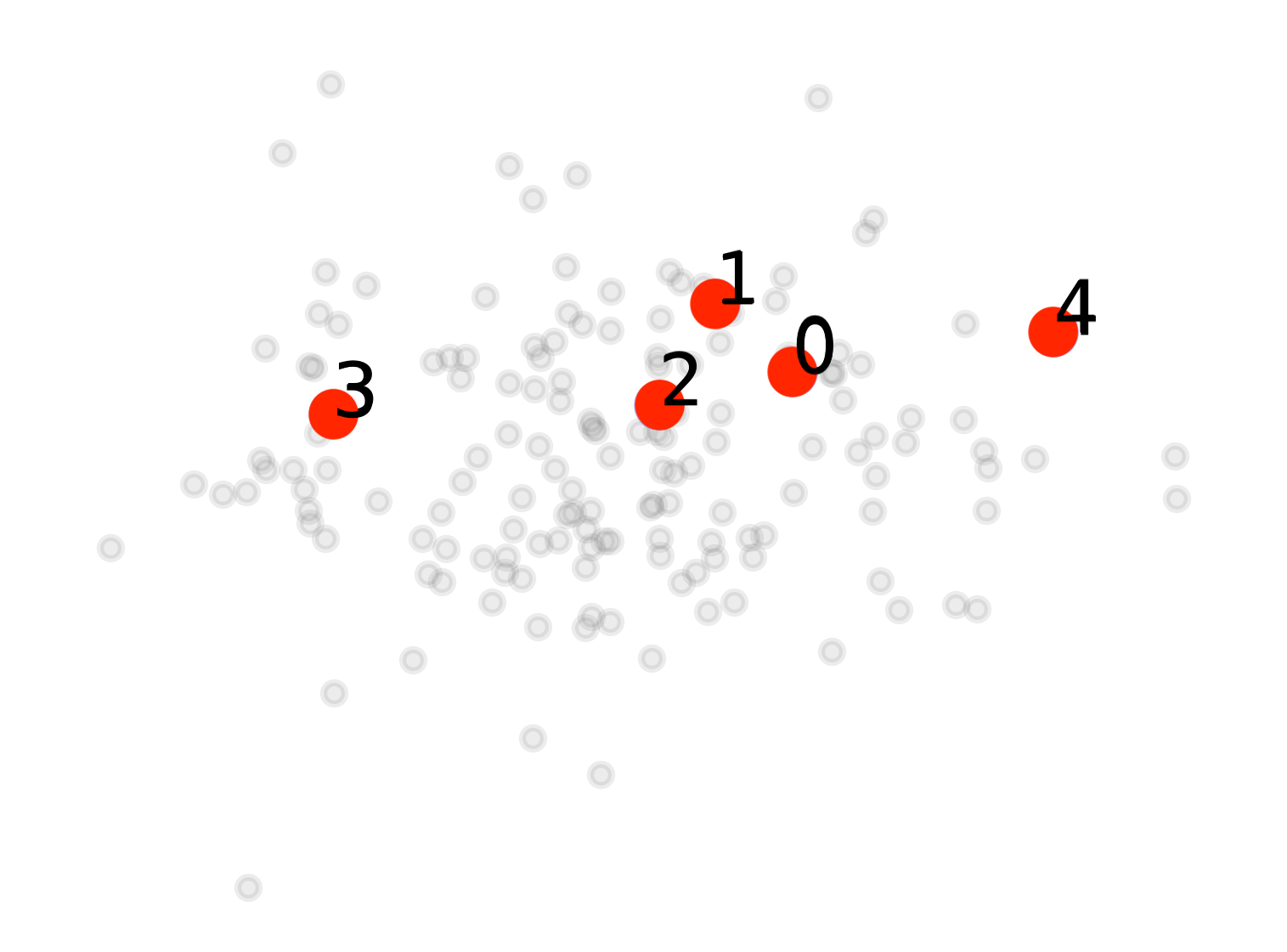}
  \end{subfigure}
  \\
\begin{subfigure}[b]{0.48\columnwidth}
    \includegraphics[width=3.6cm]{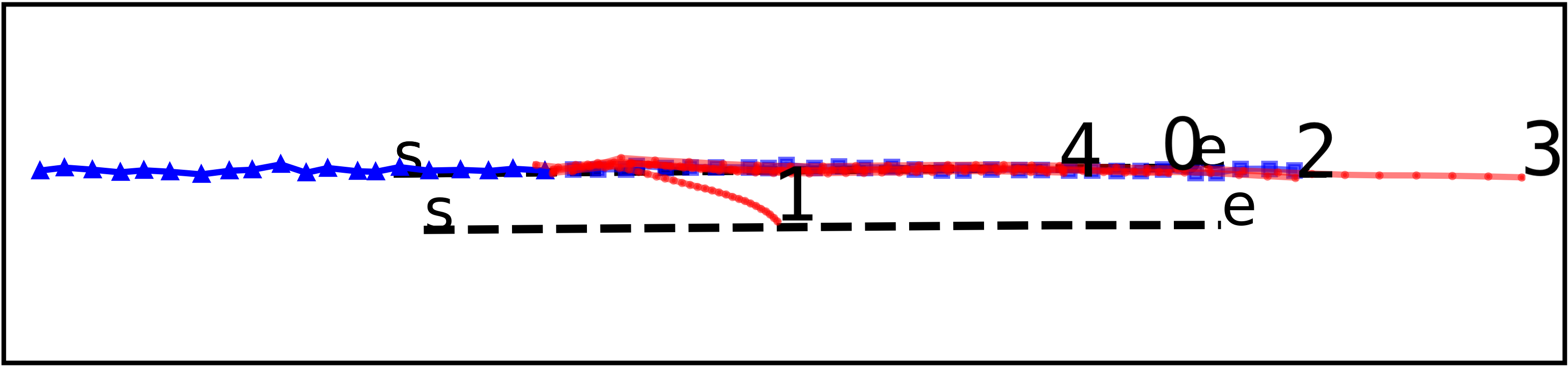}
  \end{subfigure}
  \begin{subfigure}[b]{0.48\columnwidth}
    \includegraphics[width=3.6cm]{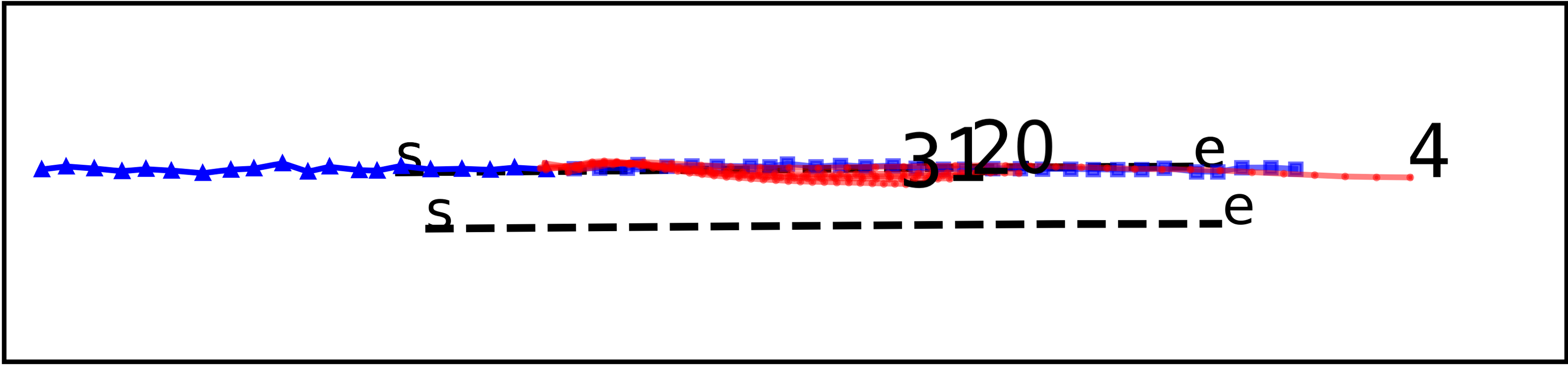}
  \end{subfigure}
  \\
   \hspace{1.3cm} FPS \hspace{2.2cm} Direct Sampling
 \\
  (b) FPS covers a low-likely lane change event that matters for decision making for the ego car.
  \caption{
  Latent space samples (numbered dots) are shown above their predicted trajectory representations. Blue: observed past and future trajectories. Red: predicted trajectory samples. Black: lane centers.
  The left column highlights samples selected by FPS and their associated predictions (numbers indicate sampling order).
  The right column shows samples using direct sampling, only covering high likelihood events.}

\label{fig:qualitative_results_sampling}
\vspace{-5mm}
\end{figure}
We show how FPS can be used to improve both prediction accuracy and diversity coverage, by illustrating two examples in Figure~\ref{fig:qualitative_results_sampling}.

\begin{figure*}[t!]
\begin{minipage}{0.08\textwidth}
FPS 
\\
\\
\\
Direct\\
Sampling
\\
\\
\end{minipage}
\begin{minipage}{0.45\textwidth}
\begin{subfigure}[b]{0.48\columnwidth}
    \includegraphics[width=3.4cm]{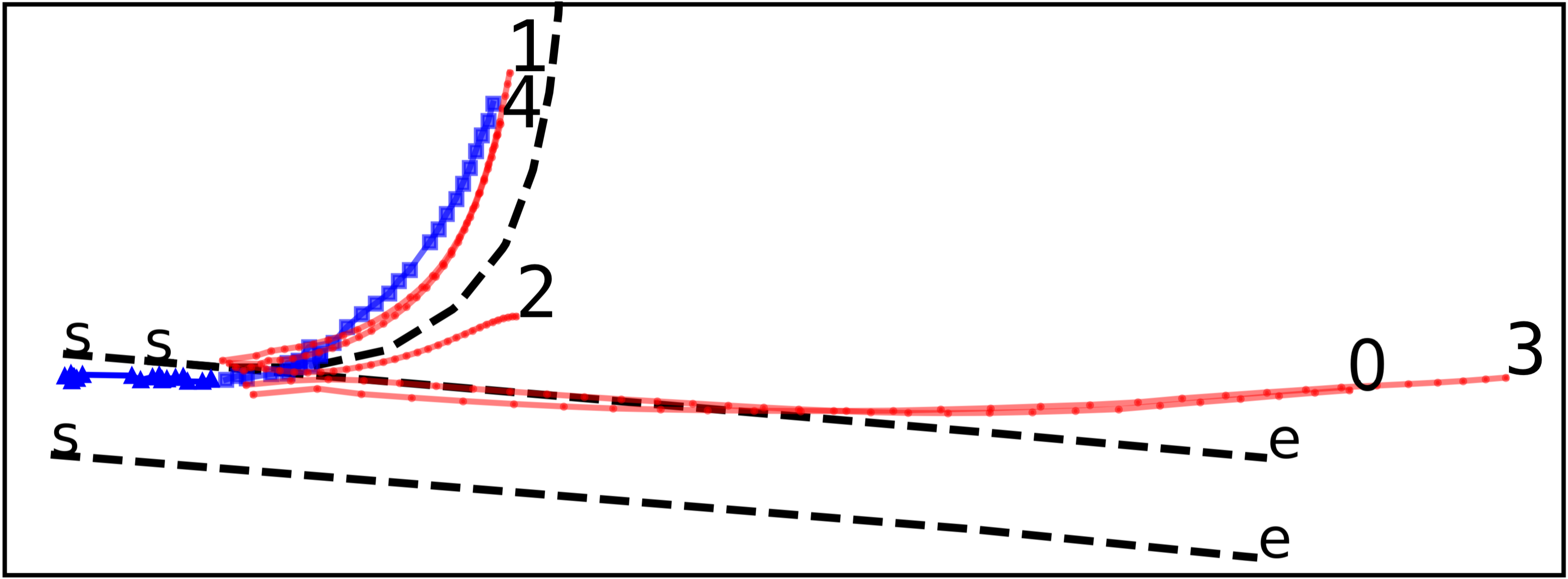}
  \end{subfigure}
  \begin{subfigure}[b]{0.48\columnwidth}
    \includegraphics[width=3.4cm]{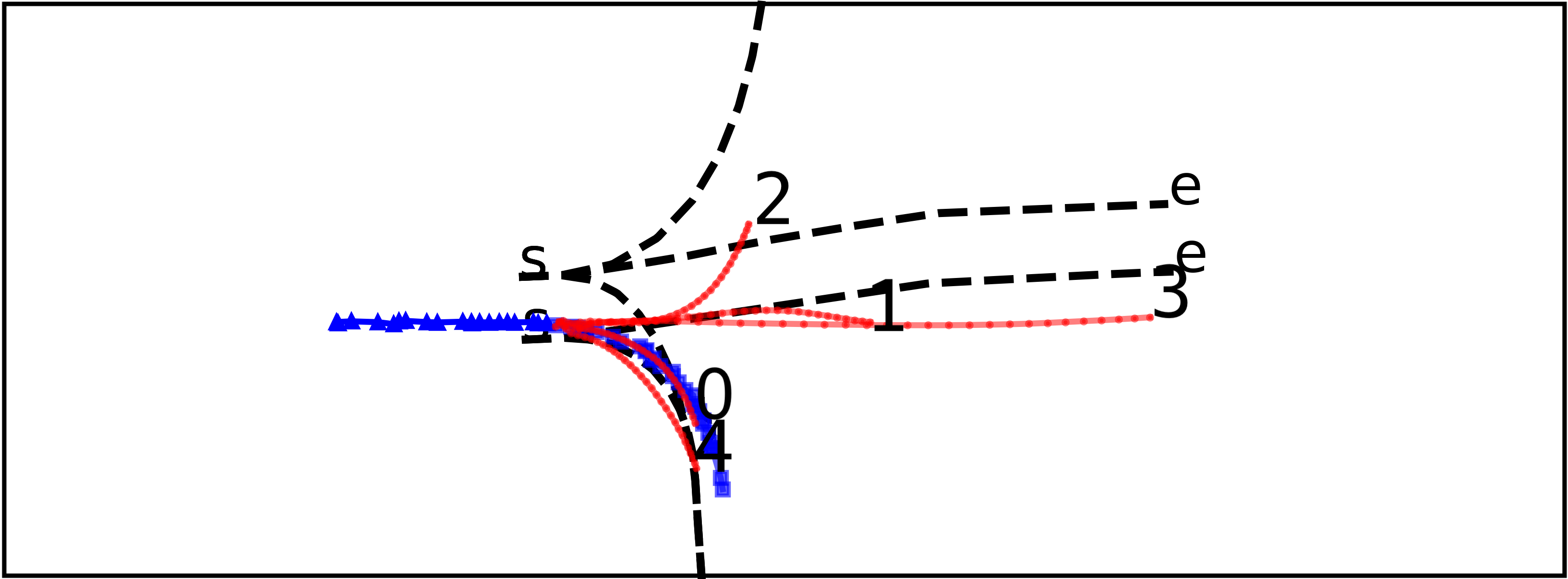}
  \end{subfigure}
  \vspace{1mm}
\\
  \begin{subfigure}[b]{0.48\columnwidth}
    \includegraphics[width=3.4cm]{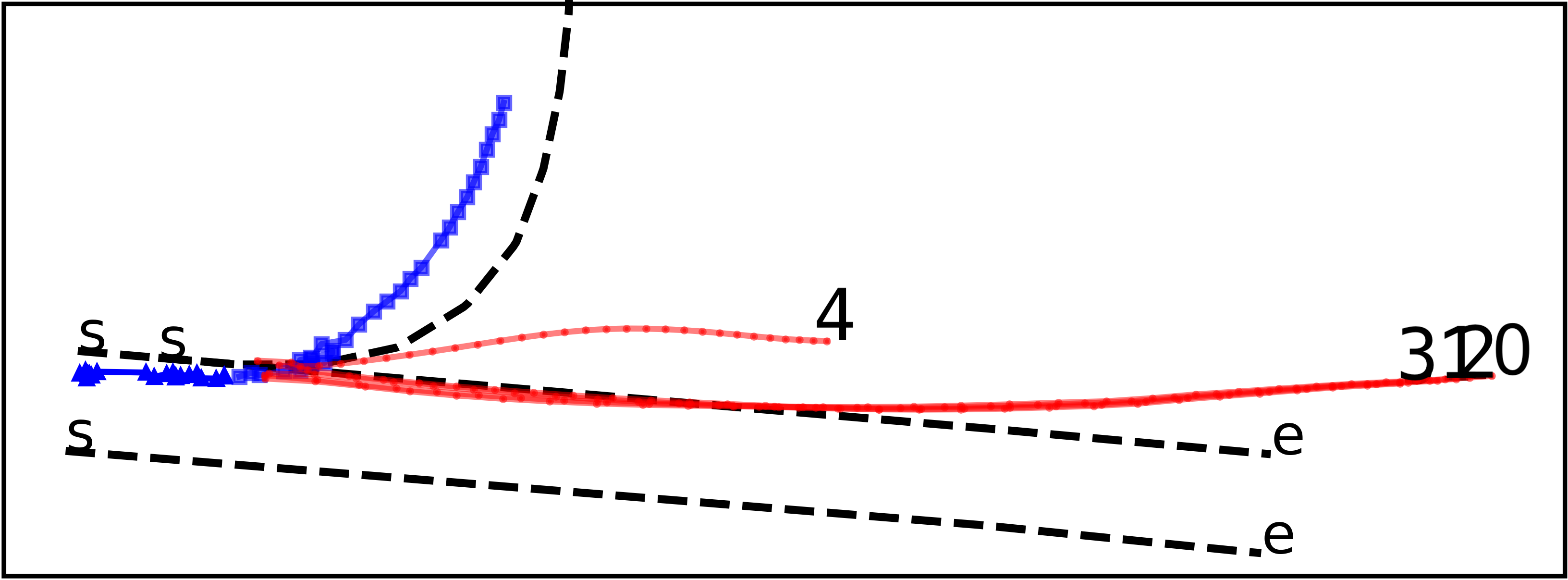}
  \end{subfigure}
  \begin{subfigure}[b]{0.48\columnwidth}
    \includegraphics[width=3.4cm]{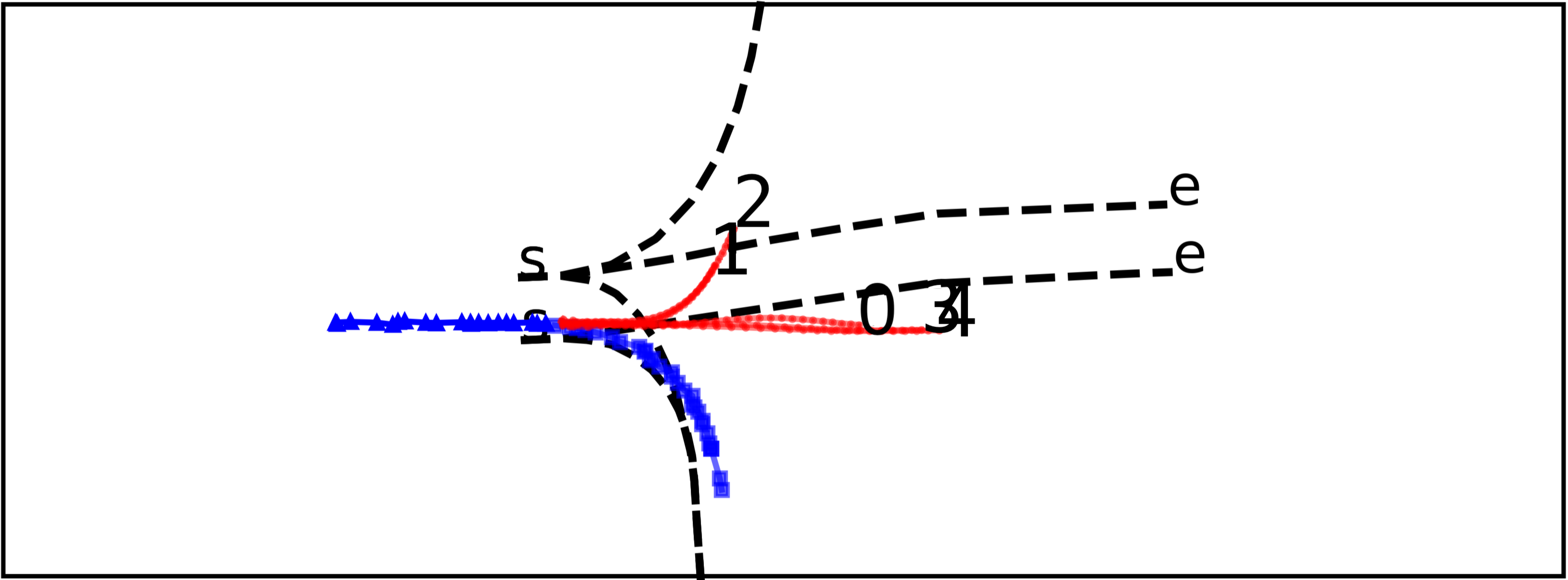}
  \end{subfigure}
  \\
  (a) Predicting diversified events helps reduce prediction error in challenging scenarios.
\end{minipage}
\begin{minipage}{0.45\textwidth}
  \begin{subfigure}[b]{0.48\columnwidth}
    \includegraphics[width=3.4cm]{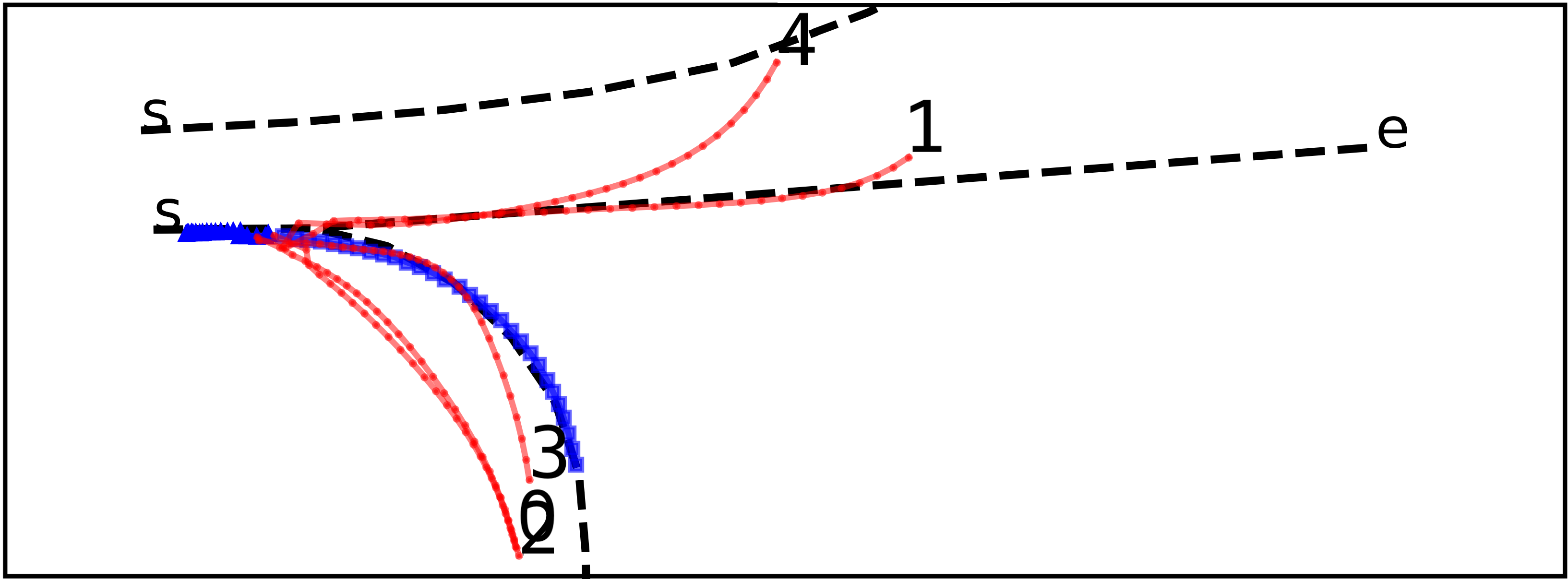}
  \end{subfigure}
  \begin{subfigure}[b]{0.48\columnwidth}
    \includegraphics[width=3.4cm]{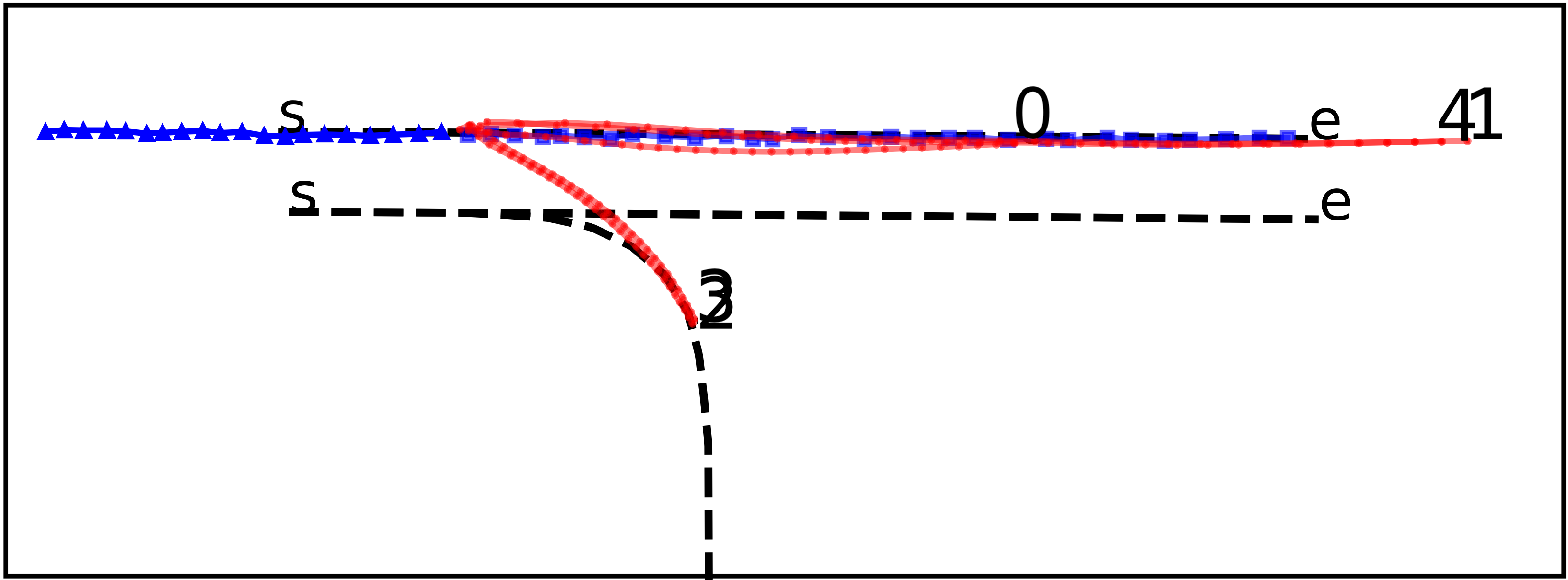}
  \end{subfigure}
  \vspace{1mm}
\\
  \begin{subfigure}[b]{0.48\columnwidth}
    \includegraphics[width=3.4cm]{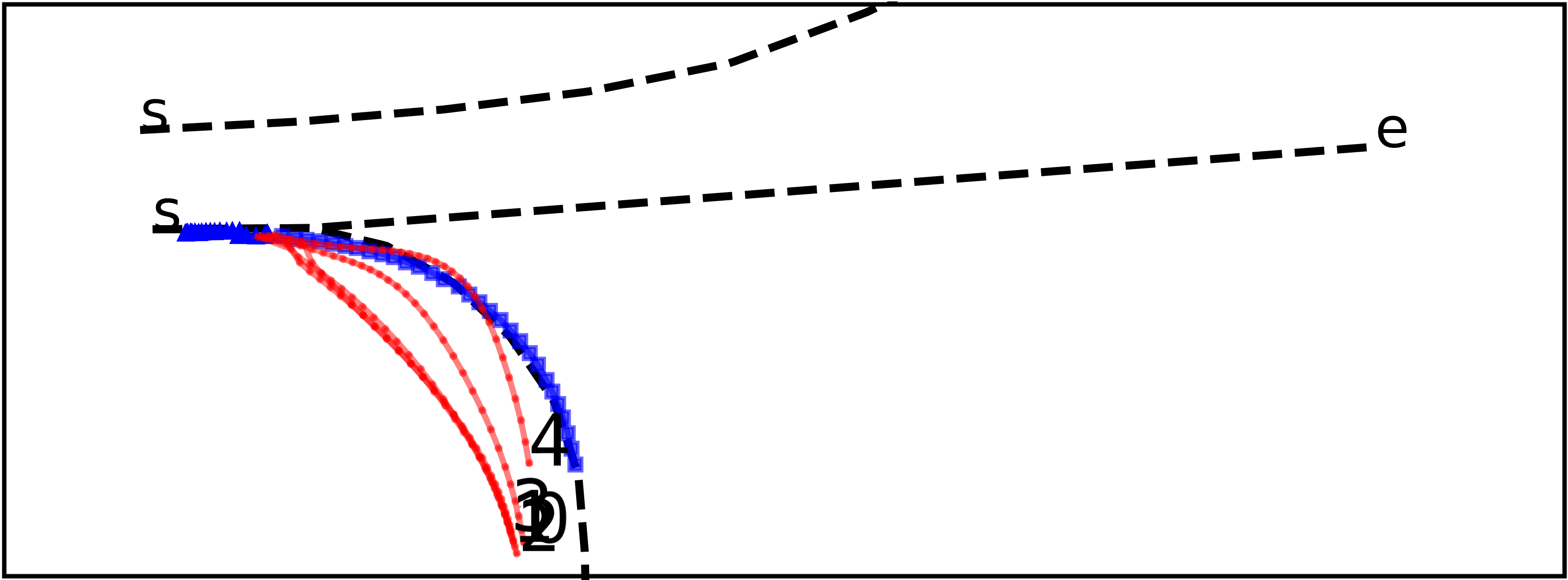}
  \end{subfigure}
  \begin{subfigure}[b]{0.48\columnwidth}
    \includegraphics[width=3.4cm]{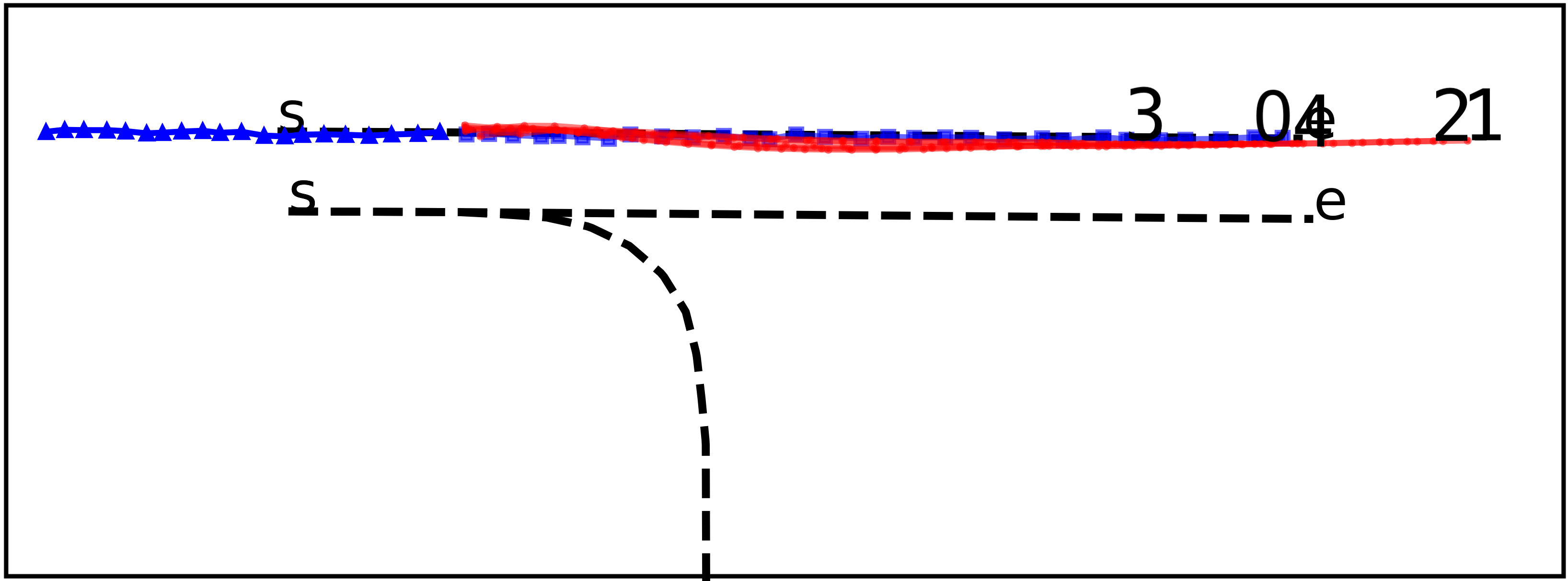}
  \end{subfigure}
  \\
  (b) Predicting merging and turning events enables robust and safe decision making for the ego car.
\end{minipage}
  \caption{Predictions of rare events in complicated driving scenarios help improve both accuracy (a) and diversity (b). Top to bottom: FPS and direct sampling with $N=5$ trajectory samples.}
  \label{fig:success_qualitative_results}
\end{figure*}

In the first example, as illustrated in Figure~\ref{fig:qualitative_results_sampling}(a), our method, as described in Algorithm~\ref{alg:sampling}, first generates $N_{all} = 200$ samples in grey, and selects $N = 5$ samples using FPS (highlighted on the left column). 
By selecting samples that are farther away, FPS is able to produce rare events, such as a right turn, as labeled in 2, which matches with the observed future trajectory and thus improves the prediction accuracy. On the other hand, direct sampling (highlighted on the right column) tends to sample points from more dense regions, which lead to high likelihood events. We show two additional challenging examples in Figure~\ref{fig:success_qualitative_results}(a), where FPS is able to reduce the prediction error, by covering turning events when the vehicle is approaching an off-ramp and a full intersection, respectively.

In the second example in Figure~\ref{fig:qualitative_results_sampling}(b), although our method predicts rare events that do not improve displacement losses compared to direct sampling, they are still important for decision making and risk estimation. Although the target vehicle is most likely to go forward, it is useful for our predictor to cover lane change behavior, as labeled in 1, even with a low likelihood, since such prediction could help avoid a possible collision if our ego car is driving in the right lane. Similarly, in other two examples, as shown in Figure~\ref{fig:success_qualitative_results}(b), our method produces events, such as merging and turning, which are unlikely to happen, but are important to consider for robust and safe decision making by the ego car.

\section{Conclusion}
\label{sec:conclusion}
We propose a vehicle motion prediction method that caters to both prediction accuracy and diversity.
We divide a latent variable into a learned semantic-level part, which embeds discrete options, and a low-level part, which encodes fine trajectory information. 
The learned geometry in the semantic part allows efficient sampling of diverse trajectories.
The method is demonstrated to achieve state-of-the-art prediction accuracy, while efficiently obtaining trajectory coverage.
Future work includes adding more complicated semantic labels such as vehicle interactions, and exploring other sampling methods beyond FPS.

\bibliographystyle{IEEEtran}
\bibliography{ref}

\end{document}